\documentclass[runningheads,a4paper]{llncs}

\usepackage{amssymb}
\usepackage{graphicx}
\usepackage{amsmath}

\usepackage{url}
\urldef{\mailsa}\path|gabriel.kronberger@fh-hagenberg.at|
\newcommand{\keywords}[1]{\par\addvspace\baselineskip
\noindent\keywordname\enspace\ignorespaces#1}

\newcommand{\gloc}{\text{loc}_g}
\newcommand{\hloc}{\text{loc}_h}
\newcommand{\hmass}{\text{mass}_h}
\newcommand{\elike}{p(e_g | \gloc, \hloc, \hmass)}

\begin{document}

\mainmatter  
\title{Declarative Modeling and Bayesian Inference of Dark Matter Halos}

\author{Gabriel Kronberger}

\authorrunning{Declarative Modeling and Bayesian Inference of Dark Matter Halos}
\institute{School of Informatics, Communications and Media,\\
University of Applied Sciences Upper Austria,\\
Softwarepark 11, 4232, Hagenberg\\
\mailsa}

\toctitle{Declarative Modeling and Bayesian Inference of Dark Matter Halos}
\tocauthor{Gabriel Kronberger}
\maketitle

\begin{abstract}
Probabilistic programming allows specification of probabilistic models
in a declarative manner. Recently, several new software systems and
languages for probabilistic programming have been developed on
the basis of newly developed and improved methods for approximate
inference in probabilistic models.  In this contribution a
probabilistic model for an idealized dark matter localization problem
is described. We first derive the probabilistic model for the
inference of dark matter locations and masses, and then show how this
model can be implemented using BUGS and Infer.NET, two software
systems for probabilistic programming. Finally, the different
capabilities of both systems are discussed. The presented dark matter
model includes mainly non-conjugate factors, thus, it is difficult to
implement this model with Infer.NET.
\end{abstract}


\section{Introduction}
% Short motivation on Bayesian modelling and probabilistic programming languages
Recently, there has been a growing interest in declarative
probabilistic modeling which has led to the development of several
software systems for probabilistic modeling and Bayesian inference
such as Stan~\cite{stan2013}, FACTORIE~\cite{factorie2009},
Infer.NET~\cite{InferNET12}, or PRISM~\cite{prism2008}. Many of these
systems provide a declarative modeling language for the definition of
probabilistic models, which allows to define random variables and
their relations in a way similar to computer programs. Thus, the term
\emph{probabilistic programming} is frequently used to refer to the
implementation of probabilistic models in such systems. The common
idea is to implement the probabilistic model declaratively, without
specifying how inference should be performed in the model. Instead the
underlying inference engine is responsible for the execution of an
appropriate inference algorithm and can potentially adapt the
inference procedure to specific models (e.g., to improve accuracy of
efficiency).

This approach to probabilistic modeling is not a recent idea;
BUGS~\cite{bugs2012}, a software system for Bayesian modeling and
inference using Gibbs sampling, is already more than twenty years old
\cite{bugs2009} and has become the de facto standard for probabilistic
programming. BUGS defines its own modeling language, and relies on the
fact, that Gibbs sampling is a very general Markov-chain Monte Carlo
(MCMC) method and allows inference in a large class of probabilistic
models. Thus, BUGS imposes almost no constraints on models and
supports a large set of models, including analytically intractable
models with non-conjugate or improper priors. However, MCMC methods
often suffer from slow convergence especially for high-dimensional
models or strongly correlated parameters. This drawback of Gibbs
sampling has been a limiting factor for probabilistic programming with
BUGS.

However, recent research results have led to several new and improved
methods for approximate inference in probabilistic models, such as
expectation propagation~(EP)~\cite{Minka2001}, variational message
passing~(VMP)~\cite{Winn2006}, and improved sampling techniques
including Hamiltonian Monte Carlo~\cite{Neal2011} and
NUTS~\cite{Hoffman2011}. Several software systems have been developed,
which incorporate these improved methods and can be used instead of
BUGS.

In this contribution we discuss a Bayesian model for an idealized
formulation of the dark matter localization problem, and show how this
model can be implemented using BUGS and Infer.NET. The aim is to
highlight and discuss the differences between BUGS and Infer.NET on
the basis of a moderately complex model. A summary of the different
capabilities of BUGS and Infer.NET, as well as a comparison of
inference results, have also been given in \cite{Wang2011}.

\subsection{Dark Matter Localization}
A large fraction of the total mass in the universe is made up of
so-called dark matter. Dark matter does not emit or absorb light but
can be detected indirectly through its gravitational field. The
existence and substance of dark matter is one of the unanswered
questions of astrophysics, and a lot of effort is spent on improving
methods to detect dark matter, and on studying its distribution in the
universe. For instance in a recent publication a map of the
distribution of dark mapper in the universe is
discussed~\cite{darkmattermap2007}.

Dark matter can be detected through the gravitational lensing effect
\cite{kaiser1993}, which occurs because the gravitational force of
large masses has a bending effect on light. Because of the
gravitational lensing effect, objects behind a mass appear displaced
to an observer and even multiple images of the same object might
appear. Additionally, the apparent shape of larger objects such as
galaxies is altered by the gravitational lensing effect. 

%% Usually a distinction between strong lensing and weak lensing effects
%% is made. The term strong lensing is used to refer to apparent
%% displacements and strong distortions of objects through gravitational
%% lensing which can also lead to multiple images. The term weak lensing
%% is used to refer to weaker distortions of the apparent shape or
%% orientation of objects such as galaxies. Dark matter halos can be
%% identified mainly through this so-called weak lensing effect because
%% the original positions and shapes of the transformed objects are
%% unknown.

The main aim of this contribution is to show, how a moderately complex
probabilistic model, such as the dark matter localization model, can
be implemented using software systems for probabilistic programming,
in order to highlight and discuss the capabilities of such
systems. We do not aim to derive a model that can be actually used for
dark matter localization. However, it should be noted that e.g.,
LENSTOOL, a software system which has actually been used for
calculating mass distribution profiles based on real images, also
implements a Bayesian model and MCMC sampling \cite{lenstool2007}.

\subsection{Synthetic Data for Dark Matter Localization}
We use a synthetic data set for the experiments presented in this
contribution as we do not aim to improve on established models for
dark matter localization. The data set has been generated for the
``Observing Dark Worlds'' competition hosted on
Kaggle\footnote{Competition organizers: David Harvey and Thomas
  Kitching, Observing Dark Worlds Competition,
  \url{http://www.kaggle.com/c/DarkWorlds/}} by simulating dark matter
halos, galaxies and the gravitational lensing effect.  Distortions
that would occur in real images e.g., through atmospheric effects or
telescopic lenses, are ignored. For the purpose of the competition,
real image data could not be used as it is necessary to compare
solutions for dark matter halo locations. The data set is composed of
300 simulated skies and either one, two, or three dark matter
halos. The data for each sky contains locations and ellipticities of
between 300 and 740 galaxies. The ellipticity is specified using two
components: the ellipticity along the x-axis $e_1$, and the
ellipticity along a 45-degree angle to the x-axis $e_2$.

\section{Model Formulation}
% General derivation and formulation of the model
In the following we describe the probabilistic graphical model for
Bayesian inference of dark matter halo locations and masses from the
observed locations and ellipticities of galaxies as specified in the
synthetic data set.

% Previous work, Tim Salimans & Probabilistic Programming for Hackers
It should be noted, that the model described below is the model that
has been used by the author for the dark worlds competition. This
model is very similar to the model used by the competition
winner\footnote{Tim Salimans' description of his model can be found at
  \url{http://timsalimans.com/observing-dark-worlds/}}, but
differs in relevant details. In particular, the model below uses
different priors.

The goal in the dark matter localization problem is to determine
probability distributions for halo locations $p(\hloc |
\gloc, e_g)$ given observed galaxy locations $\gloc =
(x,y)_h$ and ellipticities $e_g = (e_1,e_2)_g$. Using Bayes' theorem
we can derive the posterior distribution of halo locations from the
likelihood of observed locations and ellipticities times the prior.
The mass of a halo determines the strength of the gravitational
lensing effect. Therefore, the halo mass has to be included in the
model as latent variable. This leads to the following model
\begin{equation*}
p(\hloc, \hmass | \gloc, e_g) = \frac{p(\gloc, e_g | \hloc, \hmass)
  p(\hloc) p(\hmass)}{ \int_{\hmass} \int_{\hloc} p(\gloc, e_g |
  \hloc, \hmass) p(\hloc) p(\hmass)} .
\end{equation*}

There is no prior information about the locations of halos, so a flat
uniform prior $p(x_h) \sim U(0,4200), p(y_h) \sim U(0,4200)$ is used.
For the halo mass a broad gamma prior is used: $p(\hmass) \sim
\text{Gamma}(0.001, 0.001)$.

We assume that the galaxy locations are independent from halo locations
and masses
%\begin{equation}
%p(\gloc, \hloc, \text{mass}_g) = p(\gloc) p(\hloc) p(\hmass)
%\end{equation}
 so the likelihood can be transformed:
\begin{equation*}
p(\gloc, e_g | \hloc, \hmass) = p(e_g | \gloc, \hloc, \hmass) p(\gloc)
\end{equation*}

The ellipticity is given as a vector with two components $e_1, e_2$
which are independent, and the empirical distribution of ellipticities
is very close to a zero mean normal distribution with variance
$\frac{1}{20}$. Therefore, the likelihood function for observed
ellipticities $e_g$ can be expressed as a normal likelihood with the
mean given by the strength of the lensing effect and a constant
variance $\sigma^2 = \frac{1}{20}$. 
\begin{equation*}
\elike = N(e_g | f(\gloc, \hloc, \hmass), \sigma^2)
\end{equation*}
We assume that the strength of the lensing effect can be modeled using
a simple function, which increases linearly with the mass of the halo
and inversely with the distance to the halo center.
\begin{equation*}
f(\gloc, \hloc, \hmass) = \frac{\hmass}{||\gloc - \hloc ||}
\end{equation*}

\begin{figure}
\centering
\includegraphics{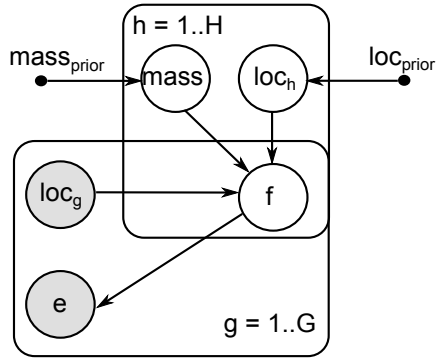}
\caption{\label{fig:prob_graph_1}Probabilistic graphical model for the
  gravitational lensing effect that can be used to infer dark matter
  locations and masses. The strength of the lensing effect $f$ is a
  result of the masses and locations of $H$ halos and effects the
  observed ellipticities of $G$ galaxies.}
\end{figure}

The final model is shown as a probabilistic graphical model in Figure
\ref{fig:prob_graph_1}, using plate notation to represent $G$ galaxies
and $H$ halos. In the graphical representation several details that
are necessary for the implementation, such as the distance of galaxies
from halos and the angle of the force vector, are not shown.

%%\begin{figure}
%%\centering \includegraphics{tangential_force}
%%\caption{\label{fig:tangential_force}The lensing effect leads to an
%%  elongation of ellipticities tangentially to the gravitational
%%  force.}
%%\end{figure}
For the implementation of the model the total force has to be
allocated to the two components of ellipticity, which describe the
elongation of the galaxy in the x-direction ($e_1$) and the elongation
along a $45\deg$ angle.  The gravitational lensing effect leads to
tangential elongation of the ellipticity. Thus, we map the tangential
force to the two components by multiplying with $-\cos(2\phi)$ and
$-\sin(2\phi)$, respectively.  $\phi$ is the angle of the vector from
the halo to the galaxy.
\begin{equation*}
%f(\gloc, \hloc, \hmass) & = \frac{\hmass}{||\gloc - \hloc ||},  \\
e_1 \sim N(- f \cos(2 \phi), \sigma^2),\: e_2 \sim N(- f \sin(2 \phi), \sigma^2),\: \phi = \text{atan}\frac{y_g - y_h}{x_g - x_h}
\end{equation*}

\subsection{Implementation using BUGS}
It is rather straightforward to transform the probabilistic model to
BUGS syntax. The model can be defined as follows: 
{\footnotesize
\begin{verbatim}
model{
  for( i in 1 : G ) {
    for( h in 1 : H ) {
      dx[i , h]    <- gx[i] - loc[h , 1]
      dy[i , h]    <- gy[i] - loc[h , 2]
      dist[i , h]  <- sqrt(dx[i , h] * dx[i , h] + 
                           dy[i , h] * dy[i , h])
      phi[i , h]   <- atan2(dy[i , h], dx[i , h])
      iDist[i , h] <- 1.0 / dist[i , h]
      f[i , h]     <- mass[h] * iDist[i , h]
      f1[i , h]    <- -f[i , h] * cos(2 * phi[i , h])
      f2[i , h]    <- -f[i , h] * sin(2 * phi[i , h])
    }
    mu1[i] <- sum(f1[i , ])
    mu2[i] <- sum(f2[i , ])
    e1[i] ~ dnorm(mu1[i], 0.05)
    e2[i] ~ dnorm(mu2[i], 0.05)
  }
  for( h in 1 : H ) {
    mass[h]    ~ dgamma(0.001, 0.001)
    loc[h , 1] ~ dunif(0, 4200)
    loc[h , 2] ~ dunif(0, 4200)
  }
}
\end{verbatim}
}

The only additional steps that are necessary to infer $\hloc$ and
$\hmass$ are loading initial values for all unobserved variables and
loading the data for galaxy ellipticities $e_1, e_2$ and locations
$\text{gx}, \text{gy}$. However, convergence is very slow when
sampling this model within BUGS.  Additionally, BUGS does not support
the $\text{atan2()}$ function in models. Fortunately, the source code
of BUGS is available, so it is possible to add support for this
function rather easily.

\subsection{Implementation using Infer.NET}
Infer.NET\footnote{Infer.NET version 2.5 is available from\\ \url{http://research.microsoft.com/en-us/um/cambridge/projects/infernet/}}~\cite{InferNET12}
allows declarative specification of probabilistic models and provides
EP~\cite{Minka2001} and VMP~\cite{Winn2006} for approximate
inference. Probabilistic models can be implemented directly in
C\#. The model is transformed transparently by the Infer.NET compiler
to C\# source code, which is then compiled to CLR byte code using the
C\# compiler. Compared to BUGS, it is much easier to use such models from
existing code, as long as the application is based on the .NET
platform. Additionally, inference is fast because the code for model
inference is compiled. The model can be implemented in the following way:
% Recently, a domain specific language \emph{Infer.NET Fun} has been
% added to Infer.NET, which allows more concise formulations of
% models. 
{\footnotesize 
\begin{verbatim}
// not shown: variable declarations
// [...]
sigma = Variable.New<double>();
evidence = Variable.Bernoulli(0.5);

IfBlock block = Variable.If(evidence);
using (Variable.ForEach(g)) {
  using (Variable.ForEach(h)) {
    // factors for the following are not implemented
    // dx[g][h].SetTo(g_x[g] - loc_x[h]);
    // dy[g][h].SetTo(g_y[g] - loc_y[h]);
    // phi.SetTo(Variable.Atan2(dy[g][h], dx[g][h]));
    // cos2phi[g][h].SetTo(Variable.Cos(2 * phi[g][h]));
    // sin2phi[g][h].SetTo(Variable.Sin(2 * phi[g][h]));
    // invDist[g][h].SetTo(1.0 / Variable.Sqrt(dx[g][h] * dx[g][h] + 
    //                                         dy[g][h] * dy[g][h]));
    f1[g][h].SetTo(-(mass[h] * invDist[g][h] * cos2phi[g][h]));
    f2[g][h].SetTo(-(mass[h] * invDist[g][h] * sin2phi[g][h]));
  }
  e1[g].SetTo(
   Variable.GaussianFromMeanAndPrecision(Variable.Sum(f1[g]), 20));
  e2[g].SetTo(
   Variable.GaussianFromMeanAndPrecision(Variable.Sum(f2[g]), 20));
}
block.CloseBlock();
\end{verbatim}
} Similarly to the BUGS implementation, arrays of random variables are
used for galaxy locations and ellipticities, as well as for halo
locations and masses. Looping over ranges can be accomplished with the
\verb+Variable.ForEach+ factor; in the example two variants to handle
blocks in Infer.NET are shown. Either the block is manually opened and
closed, as shown for the \verb+IfBlock+, or the \verb+using+-syntax is
used to manage blocks.

EP and VMP are able to exploit regularities in the model, so that
inference can be performed much faster than would be possible with
MCMC techniques. EP and VMP allow inference also for large scale
models, such as document topic modeling with latent Dirichlet
allocation. The drawback of both methods is that they are much less
general than e.g., Gibbs sampling. Thus, the set of probabilistic
models that can be used in Infer.NET is rather constrained
\cite{Wang2011}. However, in contrast to BUGS, Infer.NET also supports
conditional blocks.

Even tough EP can potentially also be used to infer marginals when the
factors are non-conjugate, this is in general not very efficient. So,
Infer.NET typically does not contain such factors. Implementing
custom factors for EP inference is rather difficult.

\begin{figure}
\centering
\includegraphics[width=5cm]{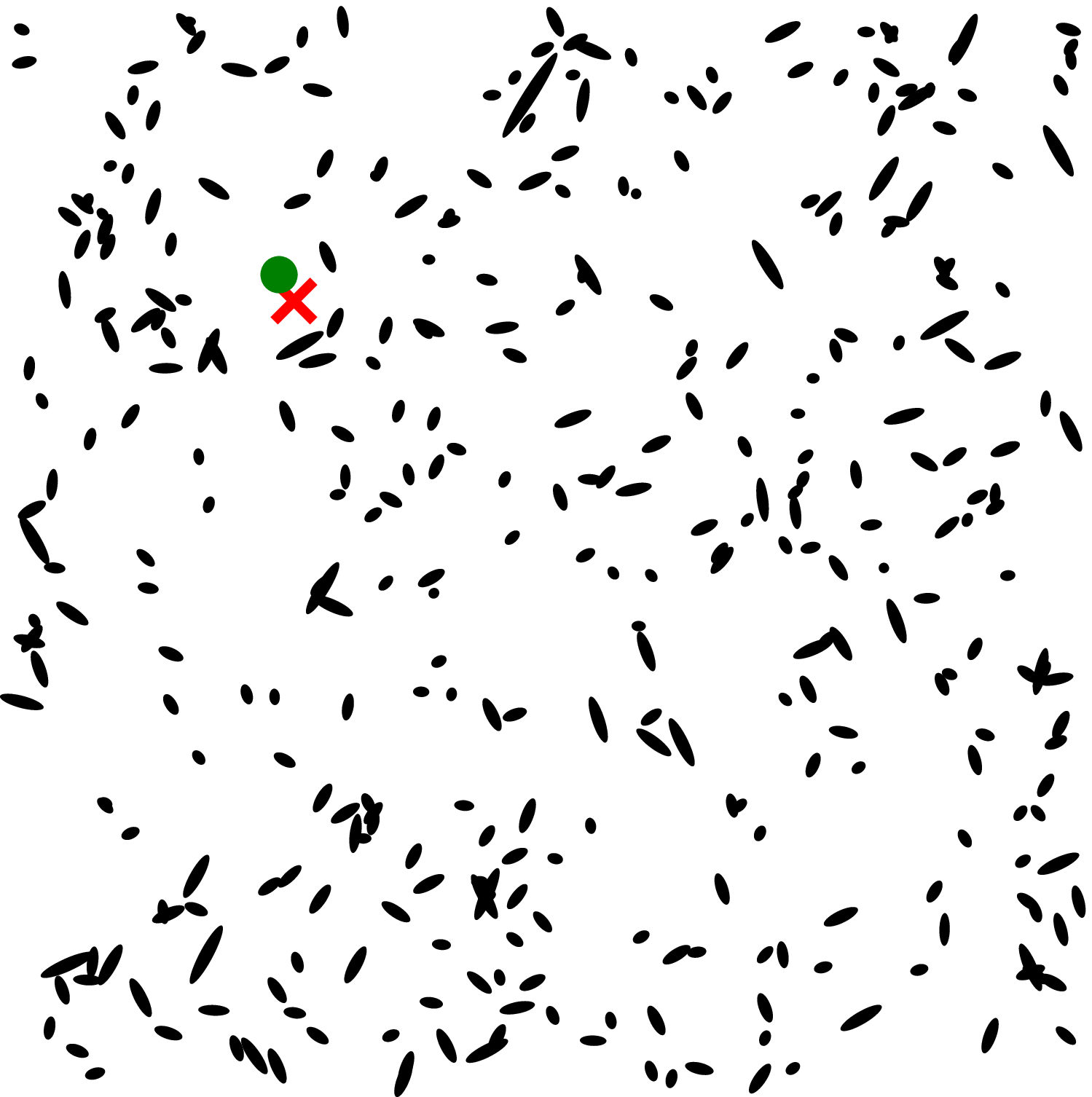}\qquad
\includegraphics[width=5cm]{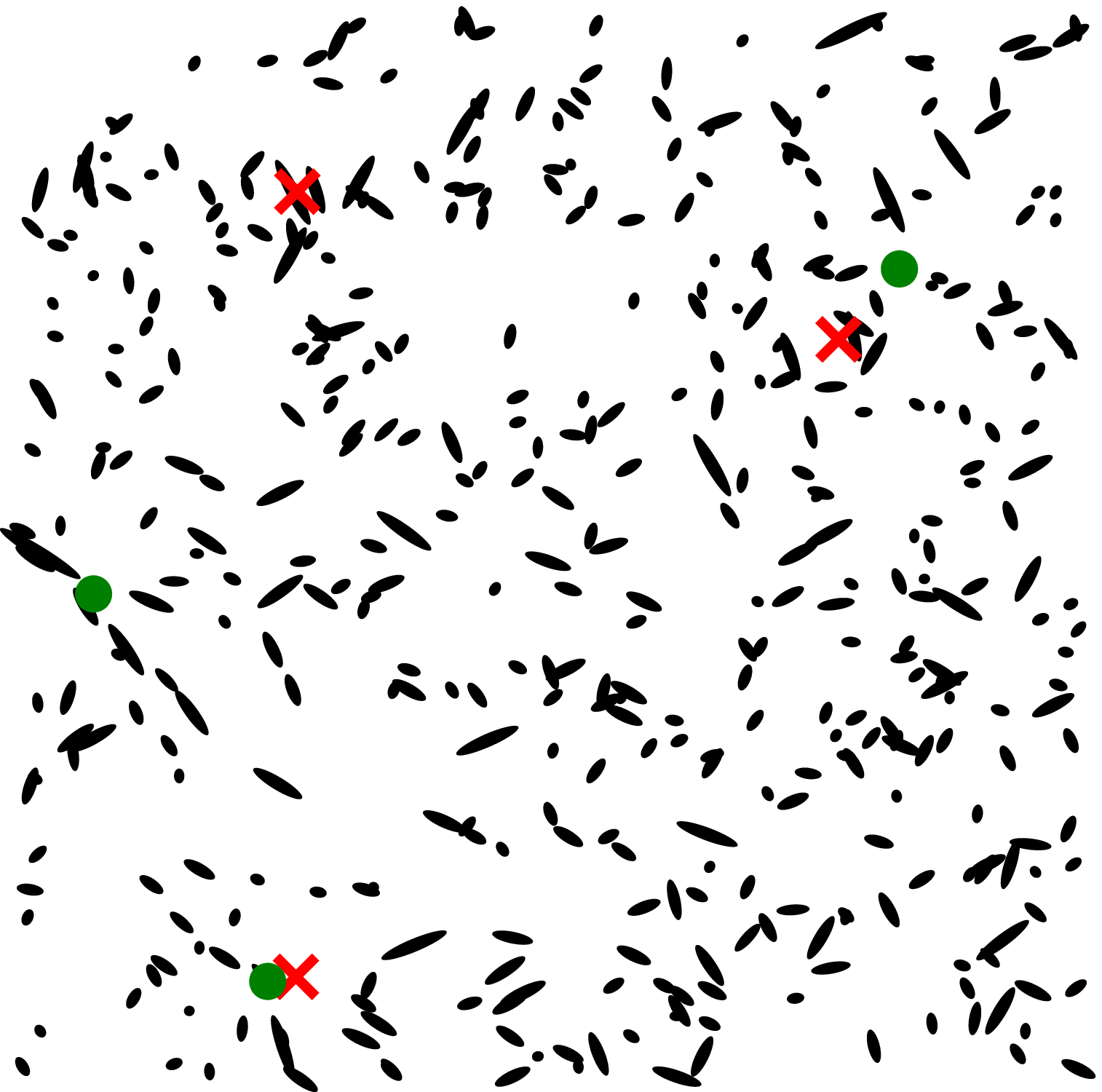}
\caption{\label{fig:likelihood} Two simulated skies showing galaxies and the actual (red cross) and predicted
  (green circle) center location of the dark matter halos.}
\end{figure}

The model uses mainly non-conjugate factors. Thus, it is not possible
to perform inference using the original model with the standard
installation of Infer.NET. It would be necessary to implement new
distribution types, factors and message operators to support inference
for this model.  Instead, the model is simplified, so that it
basically represents a simple likelihood function. Infer.NET is only
used to infer the likelihood (\verb+evidence+). The parameters of the
model, $\hloc$ and $\hmass$ are optimized w.r.t. likelihood using the
CMA-ES optimization algorithm~\cite{Hansen2001}. Figure
\ref{fig:likelihood} shows locations and shapes of galaxies and the
actual location of the dark matter halos for two skies. The predicted
halo locations are marked with a green circle.

\section{Summary and Discussion}
In this contribution a Bayesian model for gravitational lensing is
described that the author used for the ``Observing Dark Worlds''
competition. The model is similar to the winning model but uses e.g.,
different priors. The gravitational lensing model can be easily
implemented in BUGS. Only the $\text{atan2}()$ function is missing,
however, it is easy to add support for this function to the
open-source version of BUGS.  Sampling converges only slowly for this
model in BUGS.

Infer.NET uses EP or VMP for approximate inference, and the models can
be implemented directly e.g., in C\#. Inference with EP and VMP is
very efficient for certain types of models, but in general less
flexible than e.g., Gibbs sampling. Because of these restrictions, it
would be necessary to implement custom factors and distributions, to
implement the full gravitational lensing model using
Infer.NET. Instead, we performed simple ML optimization of dark halo
locations and masses, based on a simplified implementation of the
model in Infer.NET, using CMA-ES as optimization algorithm. The results
are competitive, taking a spot in the top 10\% of submitted solutions.

In this contribution we have only discussed the implementation of the
model using BUGS and Infer.NET. However, several other software
systems for probabilistic programming, e.g. Stan or FACTORIE, can also
be used instead. It would certainly be interesting to implement the
model also in these systems, and to compare the results and
performance.  Additionally, it would be interesting to discuss the
implementation of factors, which are necessary for inference with EP,
in more detail.

\end{document}